\documentclass[letterpaper, 10 pt, conference]{ieeeconf}  

\IEEEoverridecommandlockouts                              

\usepackage{amsmath}
\usepackage{amssymb}
\usepackage{mathtools}
\usepackage{graphicx}
\usepackage{float}
\usepackage{subcaption}
\usepackage{epstopdf}
\usepackage[tableposition=top]{caption}
\usepackage{array}
\usepackage{url}
\usepackage{algorithm}
\usepackage{algpseudocode}
\usepackage{diagbox}
\usepackage{multirow}
\usepackage[table]{xcolor}
\usepackage[framemethod=TikZ]{mdframed}
\usepackage[procnames]{listings}
\usepackage[hidelinks]{hyperref}

\mdfdefinestyle{frameStyle}{%
    roundcorner=5pt,
    backgroundcolor=white}

\newtheorem{definition}{Definition}

\algtext*{EndIf}
\algtext*{EndFor}
\algtext*{EndWhile}
\algtext*{EndFunction}

\newcommand{\assign}[0]{$\leftarrow$ }

\newcommand{\algin}[0]{\textbf{in} }

\title{\LARGE \bf
Ontology-Assisted Generalisation of Robot Action Execution Knowledge
}

\author{Alex Mitrevski$^{\dagger\mathsection}$, Paul G. Pl{\"o}ger$^{\dagger}$, and Gerhard Lakemeyer$^{\ddagger}$
\thanks{$^{*}$This work was supported by the b-it foundation} %
\thanks{$^{\dagger}$Alex Mitrevski and Paul G. Pl{\"o}ger are with the Department of Computer Science, Hochschule Bonn-Rhein-Sieg, Sankt Augustin, Germany\newline
        {\tt\scriptsize <aleksandar.mitrevski, paul.ploeger>@h-brs.de}} %
\thanks{$^{\ddagger}$Gerhard Lakemeyer is with the Department of Computer Science, RWTH Aachen University, Aachen, Germany
        {\tt\scriptsize gerhard@informatik.rwth-aachen.de}} %
\thanks{$^{\mathsection}$Corresponding author} %
}

\begin{document}

\maketitle
\thispagestyle{empty}
\pagestyle{empty}


\begin{abstract}

    When an autonomous robot learns how to execute actions, it is of interest to know if and when the execution policy can be generalised to variations of the learning scenarios. This can inform the robot about the necessity of additional learning, as using incomplete or unsuitable policies can lead to execution failures. Generalisation is particularly relevant when a robot has to deal with a large variety of objects and in different contexts. In this paper, we propose and analyse a strategy for generalising parameterised execution models of manipulation actions over different objects based on an object ontology. In particular, a robot transfers a known execution model to objects of related classes according to the ontology, but only if there is no other evidence that the model may be unsuitable. This allows using ontological knowledge as prior information that is then refined by the robot's own experiences. We verify our algorithm for two actions - grasping and stowing everyday objects - such that we show that the robot can deduce cases in which an existing policy can generalise to other objects and when additional execution knowledge has to be acquired.

\end{abstract}


\section{INTRODUCTION}

    When acting in everyday environments, autonomous robots need to be able to minimise the possibility of failures during execution. Failures are, however, rather likely to occur when a robot generalises its behaviour to scenarios unseen during learning, often because the robot is not even aware that its current knowledge may be insufficient or even inappropriate for performing a particular action. To deal with this problem, it is necessary to have a strategy based on which experiences of attempted generalisations are used for determining when the existing knowledge can be reused and when additional knowledge has to be acquired. This, in turn, requires a structured knowledge base that a robot can adapt as it incorporates more knowledge about its own actions.

    One well-known strategy for knowledge representation and generalisation is using an ontology \cite{beetz2018,awaad2014,alarcos2019,paulius2019}, which allows encoding knowledge about objects and more generally about environments. Ontologies include relationships between objects, which can be informative about when an action that is useful in one context can also be useful in another context.\footnote{For example, a robot that has learned how to push cups should be able to deduce that cups and glasses, which are related objects, should be pushed away in a similar way, while cups and screwdrivers, which are unrelated objects, need to be pushed away differently. An ontology can be used to encode and/or infer such knowledge.} Ontological models are, however, static in general; this makes them suitable for representing encyclopedic knowledge\footnote{An illustrative example of the extent to which encyclopedic knowledge can be specified is given in \cite{schneider2014}.}, but potentially less so for dynamically changing properties of the world. In addition, ontology-based generalisation depends on the richness of the ontology model: the more complete the ontology is, the more appropriate the behaviour of a robot using the ontology would be. For a more intelligent generalisation, the knowledge stored in the ontology needs to be augmented with the experiences of a robot itself, as this would reduce the limitations introduced by the incompleteness of the ontology model.

    \begin{figure}[tp]
        \centering
        \includegraphics[width=\linewidth]{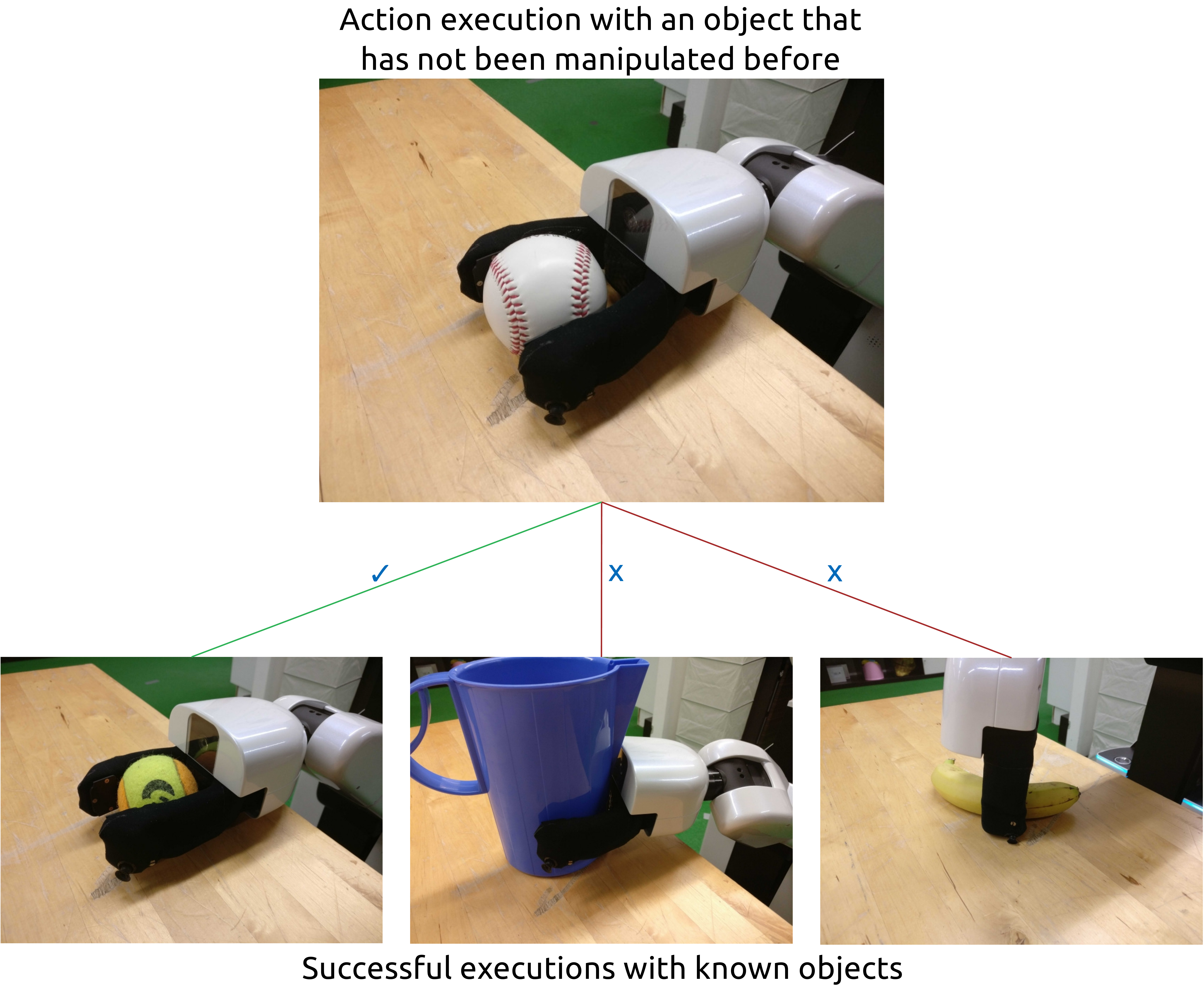}
        \caption{When generalising execution knowledge between objects, a robot should be able to consider (i) any known object relations and (ii) previous generalisation experiences that provide information about the (un)suitability of the existing execution knowledge in a given context}
        \label{fig:model_generalisation_over_objects}
    \end{figure}

    Data-driven models, particularly those based on neural networks, such as \cite{mahler2017, shao2020}, have been shown to generalise well to different, even unseen, objects; however, on its own, such generalisation is usually only based on sensory features rather than inherent knowledge about objects and their underlying contexts. While sensory features are sufficient in some applications, the applicability of learning-based methods in real-world, human-centered scenarios is generally limited, as generalisation failures are difficult to analyse without assigning a semantic meaning to the robot's decisions.

    In this paper, we address the problem of generalising action execution policies over different objects, such that we combine a plain object ontology with a probabilistic graph that indicates whether generalisation over objects is possible and likely to lead to successful execution of a particular action, such as grasping. The learned graph is derived from the object ontology, but includes edge weights that indicate the generalisation strength between object classes. We show that such a representation can be useful to determine when already learned models can be reused and when additional learning is needed. We analyse the method with learned execution models of parameterised skills as in \cite{mitrevski2020} using a Toyota HSR \cite{yamamoto2019} manipulating objects in two common domestic scenarios - grasping everyday objects and stowing them in a drawer. The results show that a robot can often reuse knowledge about objects it has manipulated before when dealing with new object classes, but also that feedback about failures can allow it to adapt its generalisation accordingly.\footnote{Accompanying video: \url{https://youtu.be/gBiYwjUTWQ8}}


\section{RELATED WORK}
\label{sec:related_work}

    Most methods that allow object generalisation are designed for a specific action, such as grasping or pushing, and are built without an ontological layer, such that generalisation is done based on extracted sensory features.
    St{\"u}ber et al. \cite{stueber2018} model a pushing action for rigid objects; the model is generalised to previously unseen objects by first using a learned density to sample contact points that most closely resemble the contacts seen in the training data and then predicting the motion of the object using a learned motion model.
    Liu et al. \cite{liu2020} develop a network-based grasping model which, given a grasp candidate and a context description, calculates the probability that the grasp candidate is suitable for the given context; the proposed architecture combines object features with contextual object and task information, which allows the method to generalise reasonably well to new objects.
    A similar grasping method is presented by Song et al. \cite{song2010}, where the action is modelled as a hybrid Bayesian network that encodes relationships between the task for which grasping is performed, object features, action parameters, and grasp constraints; by appropriate conditioning, the model can be used for object generalisation, but only to known objects.
    In principle, our method could be used in conjunction with any of these action-specific models, although we aim to reduce the data requirements of such models by leveraging prior information about objects and their relations.

    From a perceptual point of view, knowledge-assisted generalisation over objects and lifelong object learning have been addressed in different contexts.
    Denninger and Triebel \cite{denninger2018} present a lifelong learning-based object classification method which modifies the trees in a random forest as new object classes appear; a random forest is used so that new classes can be incorporated without retraining from scratch.
    In \cite{young2016}, Young et al. propose a method for inferring categories of unknown objects encountered in everyday scenes; the likely category of an unknown object is inferred using information about the object's context - represented through its surrounding objects - and querying a concept ontology, which finds the most related concept among a set of candidate concepts.
    Schoeler and W{\"o}rg{\"o}tter \cite{schoeler2016} introduce a method for recognising objects from point clouds based on affordances, which are modelled through object parts and part relations; an ontology of tools based on their functions is also defined, which allows recognising the function of unseen tools.
    As our proposed method depends on a meaningful object grounding, the above methods are complementary to ours: in particular, \cite{denninger2018} and \cite{young2016} are required for incorporating new objects, while the inclusion of affordances as in \cite{schoeler2016} would enrich the execution models of particular actions, for instance as demonstrated in \cite{gajewski2019} in the context of tool use.

    Conceptually, our work is most closely related to approaches for object and action model learning.
    Bauer et al. \cite{bauer2020} propose an approach that allows generalising actions between different objects, where the objective is to find a posterior distribution of the probability of an action effect (with a particular grounding) given a set of previous experiences of similar actions; here, actions are represented by Probabilistic Action Templates \cite{leidner2012}, where different effects are associated with probabilities, such that two instantiations are considered similar if (i) the actions use the same template and (ii) if all parameters (symbols) of the actions have a shared parent.
    Sushkov and Sammut \cite{sushkov2012} present a Bayesian active learning framework based on which a robot can identify the properties of an object or another system by testing informative actions, observing their outcomes, and updating the belief about its hypotheses of the properties; by repeating the process multiple times, the belief would converge to the correct model hypothesis.
    Sanan et al. \cite{sanan2019} describe a demonstration-based method for learning the model of an object that a robot may need to translate and/or rotate around a given axis; once such a model is learned, it can be used in a parameterised skill for manipulating the object in a particular way.
    Ivaldi et al. \cite{ivaldi2014} analyse a strategy using which a robot can acquire object models by combining perceptual features with exploratory actions, performed either by the robot itself or by human teachers.
    As in \cite{bauer2020}, we essentially generalise actions to objects that share common parents, but we aim for adaptive generalisation, since, as shown in our experiments, having a shared parent is not a sufficient condition for successful generalisation. Similar to \cite{sushkov2012}, our method explores different generalisation hypotheses, such that we aim to either converge to a valid hypothesis or conclude that more knowledge needs to be acquired. As in \cite{sanan2019}, we learn and then generalise object-specific models, but, unlike \cite{sanan2019}, we do not constrain our method to regular polyhedra. Finally, our model learning method is somewhat similar to the learning procedure in \cite{ivaldi2014}, but our main objective is to additionally learn when the learned models can be transferred between objects.


\section{BACKGROUND}
\label{sec:background}

    \subsection{Action Execution Models}

    This work builds upon \cite{mitrevski2020}, where a representation of action execution models was proposed based on which a robot jointly learns a relational model of success preconditions, or a collection of such models for multiple qualitative modes, as well as a continuous model that maps action parameters and execution constraints to predicted execution success. An execution model is learned from labelled execution data, such that the relations used for representing the relational model are defined per action, and the continuous model is represented by a Gaussian process so that prediction uncertainty can be encoded. For execution, action parameters that satisfy the relational model under a given qualitative mode are sampled from the learned success model.

    During model acquisition, a robot will typically encounter a few object classes\footnote{For instance, for the experiments in \cite{mitrevski2020}, we learned an object pulling model with only one type of object - a yogurt cup. Even such a simple model can be generalised to other object classes since there are other objects that have similar physical properties as a yogurt cup, but a robot does not necessarily know this.}, but may also need to work with other object classes when using a learned model during execution; in this case, the learned model needs to be generalised to classes other than the ones seen during training. In this paper, whenever we refer to generalisation, we mean applying a model $M_o$, which is known to be applicable for an object class $o$, to another object class $o'$.

    \subsection{Object Ontology}

    We use an ontology $\mathcal{O} = (\mathcal{T}, \mathcal{A})$ for generalisation over objects, where $\mathcal{T}$ is the TBox, which encodes object class definitions, relations between those classes, as well as object property definitions, and $\mathcal{A}$ is the ABox, which consists of class and property assertions, namely ground objects that belong to specific object classes and properties of those, respectively. For concreteness, we assume an ontology in the Web Ontology Language (OWL)\footnote{\url{https://www.w3.org/OWL/}}, although our method is not limited to OWL ontologies since we only leverage the encoded relations between object classes under the assumption that the class hierarchy represents a tree. We only consider generalisation over the TBox, although the presented algorithm can be extended to the ABox as well, but at an expected higher computational cost.


\section{EXECUTION MODEL GENERALISATION}
\label{sec:execution_model_generalisation}

    For generalising execution models between objects intelligently, a robot that needs to perform an action with a given object, such as pulling the object to a target region, needs to be aware of whether it has manipulated that particular object or similar objects before. If an explicit execution policy for an object class is not known, the robot has to decide how it could generalise its existing knowledge to the new situation.

    In this paper, we propose a method for ontology-assisted execution model generalisation that involves two concepts: (i) based on the ontology, a \emph{subset of related object classes} is identified, namely classes whose execution models (if already known) are useful to consider when executing an action with an object for which there is no known execution model, and (ii) a probabilistic model is created for the related classes, which indicates whether an execution model learned for a given action and a particular object class is \emph{suitable} when used for executing the same action for another object class.

    The objective of the subset of related classes is twofold: first of all, it constrains the search for execution models that may be reused to a computationally manageable level, but more importantly, it prevents generalisation between object classes that are unrelated according to the ontology model, thereby reducing the possibility of execution failures due to inappropriate generalisation.\footnote{Our assumption here is that related object classes have similar physical characteristics, which is useful prior knowledge for a robot to have. Our method is still applicable if this does not hold for a given ontology, but a robot may need to perform more generalisation trials in that case.} The probabilistic model also has a dual purpose: it weighs the relations encoded by the ontology in order to reinforce meaningful generalisations and allows experience-based acquisition and improvement based on the resulting outcomes of attempted generalisations.

    \subsection{Model Selection for Generalisation}

\begin{figure}[tp]
        \begin{subfigure}[t]{0.475\linewidth}
            \centering
            \includegraphics[scale=0.13]{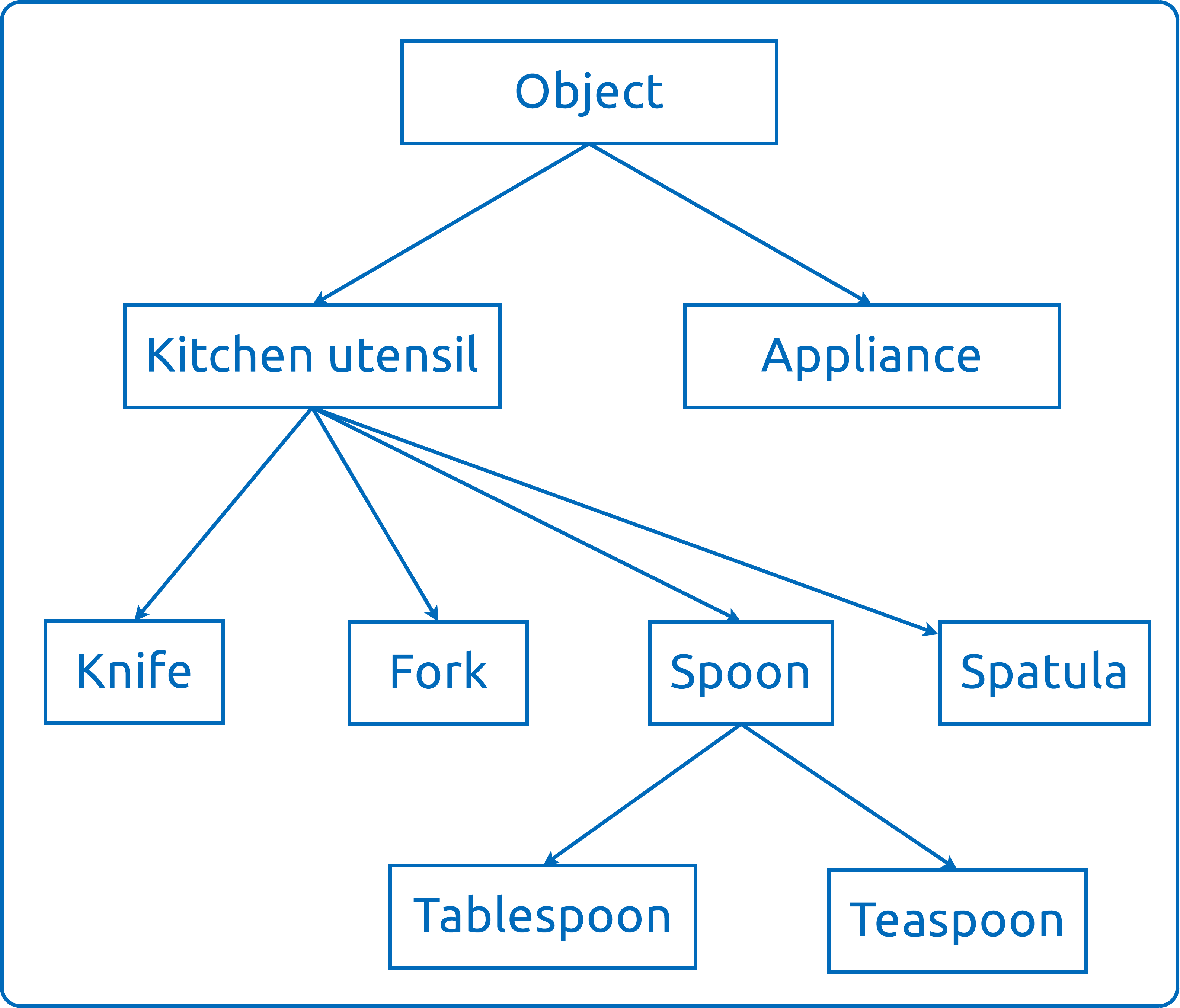}
            \caption{Example class hierarchy in an ontology}
        \end{subfigure}
        \hspace{0.025\linewidth}
        \begin{subfigure}[t]{0.475\linewidth}
            \centering
            \includegraphics[scale=0.13]{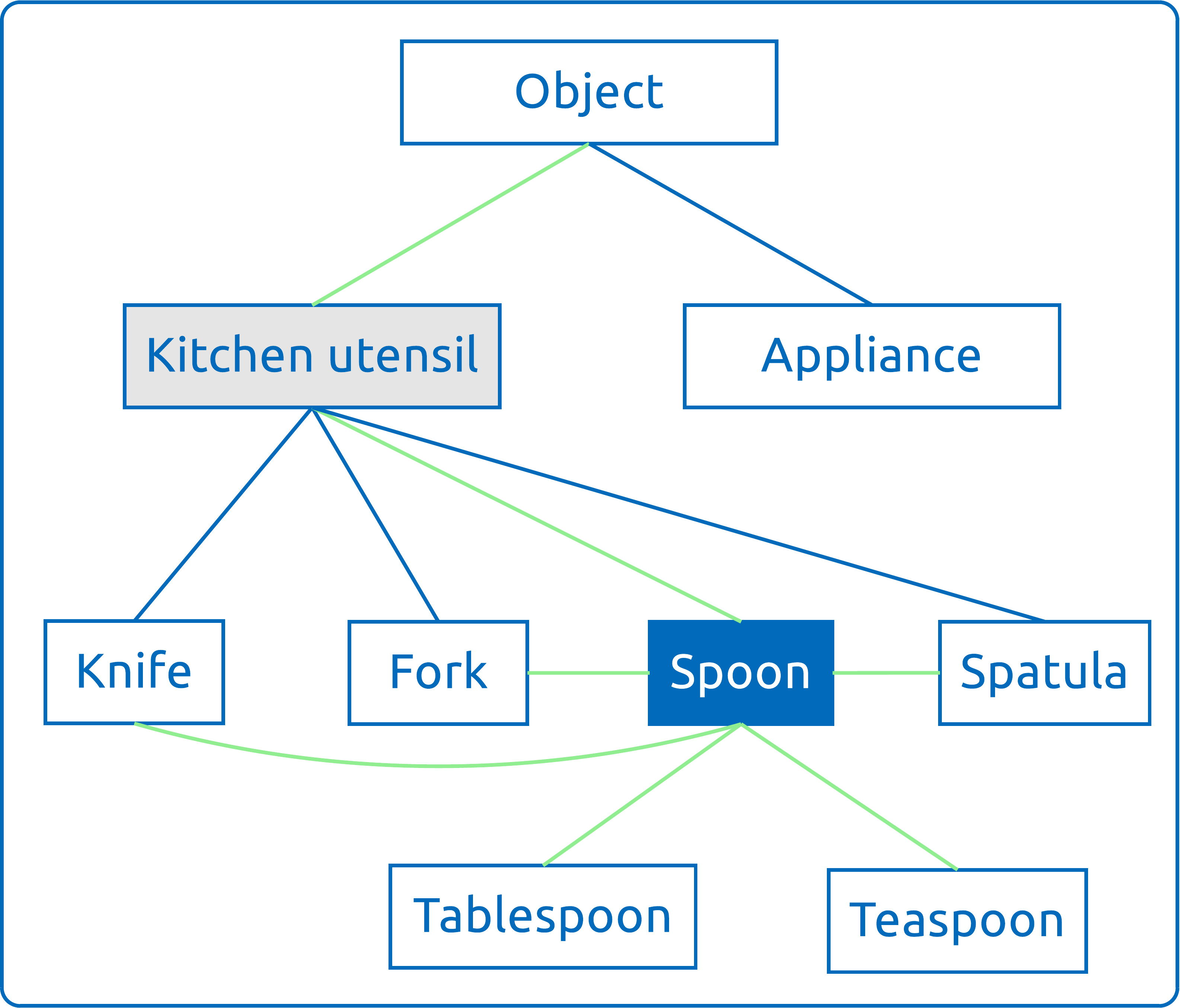}
            \caption{Induced object cluster for a given object}
        \end{subfigure}
        \caption{Illustration of an object cluster in an ontology. The blue node is the object class $o$ whose cluster $C_o$ is being extracted. The green edges indicate object classes that could be considered for generalisation if an execution model for $o$ is not known. Gray nodes are classes for which an execution model is not known, and thus do not belong to $C_o$.}
        \label{fig:ontology_and_suitability_graph}
    \end{figure}

    Our proposed method for ontology-based generalisation is based on the idea of what we call a \emph{suitability graph}, which encodes the relatedness and suitability concepts discussed above. A suitability graph is derived from an object cluster, which is defined separately for each object class $o$.
    \begin{mdframed}[style=frameStyle]
        \begin{definition}
            Given an ontology $\mathcal{O}$ and an object class $o$, an \emph{object cluster} $C_o$ is the set of ancestor, sibling, and children classes of $o$ in $\mathcal{O}$, such that there is an execution model $M_{\tilde{o}}$ for every $\tilde{o} \in C_o$.
        \end{definition}
    \end{mdframed}
    A cluster $C_o$ represents a set of direct and indirect relations between classes in an ontology, namely an object class $o$ is directly related to its parents and children, and indirectly to its siblings and ancestors. The reasoning behind this definition is as follows: (i) considering ancestor classes allows knowledge transfer from a general case to a specific case, (ii) children classes are exactly the opposite, as knowledge could be transferred from a specific case to a more general case, and (iii) since sibling classes have shared parents, they may also transfer knowledge between each other. Fig. \ref{fig:ontology_and_suitability_graph} illustrates the notion of an object cluster for kitchen objects.

    Given $C_{o}$, we know which object classes may be considered for generalisation during execution, but not how appropriate each individual generalisation would be. We represent the probability that the model of an object class $\tilde{o}$ is selected for executing an action $a$ under a qualitative mode $q$ by a weighted probability distribution of the form
    \begin{equation}
        P_{t+1}(\tilde{o}|o,S) = \eta \operatorname{s}(o, \tilde{o})P(S|\tilde{o},o)P_{t}(\tilde{o}|o,S)
        \label{eq:model_distribution}
    \end{equation}
    where $\operatorname{s}(o,\tilde{o})$ represents the similarity between $o$ and $\tilde{o}$ in the ontology, $P(S|\tilde{o},o)$ is the probability that selecting the model of $\tilde{o}$ leads to successful execution ($S$ is thus a Bernoulli variable), $\eta$ is a normalising constant, and $P_{t}(\tilde{o}|o,S)$ is a recursively updated value, such that $P_{0}(\tilde{o}|o,S)$ is uniform. It should be noted that all probabilities above are additionally conditioned on an action $a$ and a qualitative mode $q$, but we have omitted these for simplifying the notation. This distribution defines a probabilistic model that weighs the relations between the classes in $C_{o}$; we call this weight suitability, and the model itself a suitability graph.
    \begin{mdframed}[style=frameStyle]
        \begin{definition}
            For an object class $o$ and an action $a$, which is potentially executed under a qualitative mode $q$, a \emph{suitability graph} $\mathcal{G}_{o,a}$ models a distribution $P(\tilde{o}|o,S,a,q)$ for every $\tilde{o} \in C_o$, which represents the probability that using the execution model of $\tilde{o}$ to execute $a$ with $o$ will result in successful execution.
        \end{definition}
    \end{mdframed}

    Suitabilities are composed of two major components - the success probability distribution $P(S|\tilde{o},o)$ and the similarity coefficient $\operatorname{s}(o, \tilde{o})$ - which are described below.

    \paragraph{Success probability distribution} We model the problem of estimating $P(S|\tilde{o}, o)$ as a sequence of Bernoulli trials with an unknown success probability. Similar to \cite{bauer2020}, the success probability is found by estimating the parameters of a beta distribution $\operatorname{Beta}(\alpha_{o_{\tilde{o}}}, \beta_{o_{\tilde{o}}})$, whose density is given as
    \begin{equation}
        p(\theta) = \gamma \theta^{\alpha_{o_{\tilde{o}}}-1}(1 - \theta)^{\beta_{o_{\tilde{o}}}-1}
    \end{equation}
    where $\gamma$ is a normalising constant. Given $\alpha_{o_{\tilde{o}}}$ and $\beta_{o_{\tilde{o}}}$, we use $P(S=1|\tilde{o}, o) \sim \operatorname{Beta}(\alpha_{o_{\tilde{o}}}, \beta_{o_{\tilde{o}}})$ to represent the success probability when applying $\tilde{o}$'s model to $o$. To indicate the robot's initial ignorance about the suitability of $\tilde{o}$'s model, $P(S=1|\tilde{o}, o) \sim \operatorname{Beta}(\alpha_0, \beta_0)$ before any attempted generalisations, where $\alpha_0 = \beta_0$. The parameters of the beta distribution are subsequently modified according to the results of the action executions. In particular, following \cite{koller2009} and denoting by $N_{o,\tilde{o}}$ the number of times the action has been performed for object class $o$ using the model of $\tilde{o}$, the posterior parameters of the beta distribution are given as
    \begin{equation}
        \operatorname{Beta}(\alpha_{o_{\tilde{o}}}, \beta_{o_{\tilde{o}}}) = \operatorname{Beta}\left(\alpha_0 + N_{o,\tilde{o}}^{+} - 1, \beta_0 + N_{o,\tilde{o}}^{-} - 1\right)
        \label{eq:success_distribution}
    \end{equation}
    where $N_{o,\tilde{o}}^{+} = \sum_{i}X^{o}_{\tilde{o},i}$ with $X^{o}_{\tilde{o},i} = 1$ if the $i$-th execution was successful and $0$ otherwise, and $N_{o,\tilde{o}}^{-} = N_{o,\tilde{o}} - N_{o,\tilde{o}}^{+}$. In practice, to estimate $P(S=1|\tilde{o}, o)$, we draw a collection of samples $\mathcal{B}$ from $\operatorname{Beta}(\alpha_{o_{\tilde{o}}}, \beta_{o_{\tilde{o}}})$ rather than a single sample and then set $P(S=1|\tilde{o}, o) = \frac{\sum_{i}\mathcal{B}_i}{|\mathcal{B}|}$.\footnote{We use a mean estimator instead of the expected value $\frac{\alpha}{\alpha + \beta}$ to encourage small random exploration over close posterior values in Eq. \ref{eq:model_distribution}.}

    \paragraph{Object similarity} To find the similarity between two classes in an ontology, we use the Wu-Palmer (WUP) similarity introduced in \cite{wu1994}:
    \begin{equation}
        \operatorname{s}(o,\tilde{o}) = 2\frac{\operatorname{depth}\left(\operatorname{LCS}(o, \tilde{o})\right)}{\operatorname{depth}(o) + \operatorname{depth}(\tilde{o})}
    \end{equation}
    where $\operatorname{depth}: O \rightarrow \mathbb{N}$ is the depth of an object class in the hierarchy induced by the ontology and $\operatorname{LCS}: O \times O \rightarrow O$ is the least common subsumer of $o$ and $\tilde{o}$, namely the most specific class that is a common ancestor of $o$ and $\tilde{o}$.\footnote{$\operatorname{s}(o,\tilde{o})$ ranges between $0$ and $1$, with $\operatorname{s}(o,\tilde{o}) = 1$ when $o = \tilde{o}$.} The WUP similarity corresponds well to $\mathcal{G}$ as it considers (i) siblings to be more similar to each other than nodes among different levels of the hierarchy, which puts an intuitive prior preference on generalisation between sibling classes and (ii) classes with higher depths in the hierarchy to have higher similarity to each other than classes near the top of the hierarchy, which encodes the principle that class similarity increases with the specificity of the ontology model.

    \paragraph{Model selection} Given the posterior values $P_{t+1}(\tilde{o}|o,S)$ for each $\tilde{o} \in C_o$, an execution model $M_{o^{*}}$ is selected for the object class $o^{*}$ that maximises the posterior:
    \begin{equation}
        o^{*} = \underset{\tilde{o} \in C_o}{\arg\max} \, P_{t+1}(\tilde{o}|o,S=1)
    \end{equation}
    If there are multiple models that maximise the posterior, all of them are considered to be equally applicable, so one of the models is chosen at random.\footnote{If $|C_o| = 1$, the posterior of the only related object will always be $1$.} The model selection process is summarised in Alg. \ref{alg:executionModelSelection}.
    \begin{algorithm}[tp]
        \begin{algorithmic}[1]
            \Function{\texttt{generaliseExecutionModel}}{$o$, $X^{o}$, $t$}
                \State $M_o$ \assign \texttt{getModel}($o$)
                \If{$M_o \neq \varnothing$}
                    \State \texttt{executeAction}($M_o$, $o$)
                    \State \Return
                \EndIf
                \State $C_o$ \assign \texttt{getObjectCluster}($o$)
                \For{$\tilde{o}$ \algin $C_o$}
                    \State $P_{t+1}(\tilde{o}|o,S) = \eta \operatorname{s}(o, \tilde{o})P(S|\tilde{o},o)P_{t}(\tilde{o}|o,S)$
                \EndFor
                \State $o^{*}$ \assign $\underset{\tilde{o} \in C_o}{\arg\max} \, P_{t+1}(\tilde{o}|o,S=1)$
                \State $M_{o^{*}}$ \assign \texttt{getModel}($o^{*}$)
                \State $outcome$ \assign \texttt{executeAction}($M_{o^{*}}$, $o$)
                \State $X^{o}_{o^{*}}$ \assign $X^{o}_{o^{*}} \cup outcome$
            \EndFunction
        \end{algorithmic}
        \caption{Execution Model Selection and Action Execution With the Selected Model}
        \label{alg:executionModelSelection}
    \end{algorithm}

    Let us now briefly consider how the distribution in Eq. \ref{eq:model_distribution} will evolve over time. Given an object class $o$ and before incorporating any evidence about attempted generalisations, classes closer to $o$ in $\mathcal{O}$ will clearly be preferred for generalisation, as the expression is dominated by $\operatorname{s}(o,\tilde{o})$; however, as the robot attempts to generalise its knowledge throughout its lifetime, its own experiences will start having a more prominent role in the distribution. This will counteract potentially misleading information encoded in the ontology, thereby allowing a robot to exhibit lifelong learning capabilities. Our experiments demonstrate this for some of the objects for which multiple candidate models were available.\footnote{One thing to note is that $\operatorname{s}(o,\tilde{o})$ could, in principle, be incorporated in Eq. \ref{eq:model_distribution} through the initial parameters of the beta distribution, but keeping it as an outside multiplicative factor makes it easier to incorporate changes in the ontology hierarchy, as those would only be reflected in the value of $\operatorname{s}(o,\tilde{o})$ without disrupting previously collected generalisation experiences.}

    \subsection{Model Generality and New Model Acquisition}

    As models are generalised among objects throughout a robot's lifetime, the robot will acquire knowledge about their applicability for different object classes. This may lead to different outcomes regarding the extent to which a model is applicable: (i) a model can be reliably generalised to all sibling classes, and hence to a parent class, or (ii) existing models are not appropriate for a specific object class, so a robot needs to perform additional learning experiments to acquire a new model for that class. We propose two heuristics to control the generalisation and specification of models.
    \begin{mdframed}[style=frameStyle]
        \emph{Generalisation heuristic}: A model $M_{o}$ is general enough to be transferred to a parent class if, for every sibling class $\tilde{o}$ of $o$, $P(S=1|\tilde{o}, o) \geq \tau$, where $\tau$ is a predefined certainty threshold.
    \end{mdframed}
    \begin{mdframed}[style=frameStyle]
        \emph{Specification heuristic}: A new model $M_{o}$ for an object class $o$ has to be learned if either $|C_o|=0$ or, $\forall\tilde{o} \in C_{o}$, $P(S=0|\tilde{o}, o) \geq \tau$.
    \end{mdframed}
    Both of these heuristics have a single hyperparameter: the certainty threshold $\tau$. The definition in Eq. \ref{eq:success_distribution} guarantees that the belief of $S=1$ and $S=0$ will increase smoothly with the number of execution successes and failures respectively, such that a reasonably high $\tau$ will prevent premature decisions about the (un)suitability of a model. For instance, using $\alpha_0 = \beta_0 = 1$ and $\tau = 0.8$ means that a model $M_{o}$ can be generalised to a parent class if at least three successful and no failed executions have been observed for all sibling classes, but at least eight successful executions will be required if at least one failed execution is observed. This behaviour is rather intuitive and desirable: the more inconclusive the generalisation results are, the more investigation a robot needs to do before drawing conclusions about the need for generalising or specifying a model.


\section{EXPERIMENTS}
\label{sec:experiments}

\begin{figure}[tp]
        \begin{subfigure}[t]{0.475\linewidth}
            \centering
            \includegraphics[width=\linewidth]{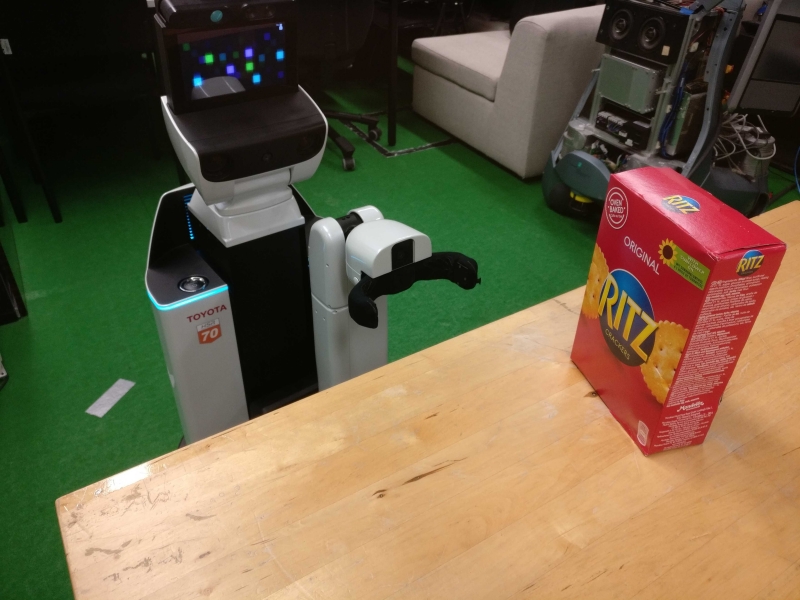}
            \caption{Grasping an object}
        \end{subfigure}
        \hspace{0.025\linewidth}
        \begin{subfigure}[t]{0.475\linewidth}
            \centering
            \includegraphics[width=\linewidth]{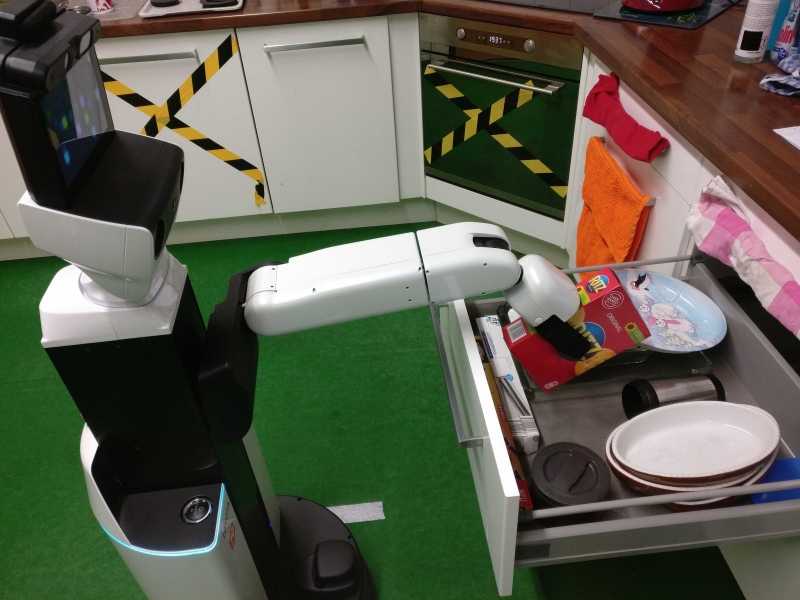}
            \caption{Object stowing in a drawer}
        \end{subfigure}
        \caption{Illustration of the setup for the experimental use cases}
        \label{fig:experimental_setup}
    \end{figure}

    We evaluate our method for ontology-assisted generalisation by considering various domestic objects in the context of two actions performed by a Toyota HSR: grasping an object for subsequent transportation and stowing a grasped object in a drawer. The experimental setup is illustrated in Fig. \ref{fig:experimental_setup}.

\begin{figure}[tp]
        \centering
        \includegraphics[width=\linewidth]{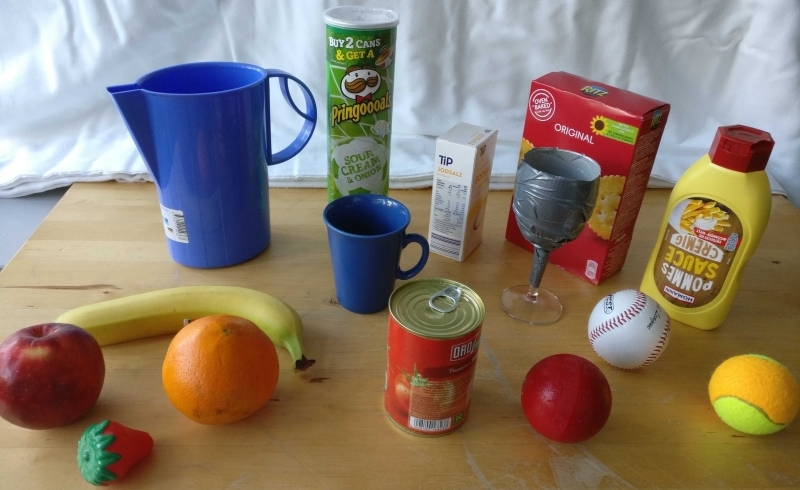}
        \caption{Objects used in the experiments}
        \label{fig:experiment_objects}
    \end{figure}

    For the experiments, we use a subset of object classes from the YCB object set \cite{calli2015}, shown in Fig. \ref{fig:experiment_objects}, in particular fruits (banana, apple, orange, strawberry), food and drink containers (chips can, tomato can, cracker box, sugar box, mustard container, mug, wine glass, pitcher), and balls (tennis ball, baseball, racquetball).\footnote{As we do not have access to the actual YCB objects, we use local objects that closely resemble the YCB objects.}
    We use an OWL ontology similar to KnowRob \cite{beetz2018} for organising the object classes, such that most YCB classes and other common domestic objects are included in the ontology\footnote{The ontology used in the experiments can be found at \url{https://github.com/b-it-bots/mas_knowledge_base/blob/devel/common/ontology/apartment.owl}}; we do not directly use KnowRob because some of the YCB classes are not included there.

    For both actions, we learn dedicated execution models (using guided learning as in \cite{mitrevski2020}) for each of the following objects in order to have a reasonable variety of models: apple, chips can, sugar box, mug, and tennis ball. Each model is learned using $25$ executions of the action; the remaining objects are used for testing the generalisation. For testing, we perform $10$ trials for each action and test object, such that $P(S|\tilde{o}, o)$ is updated after every execution based on the outcome and $P_{t}(\tilde{o}|o,S)$ is updated according to Eq. \ref{eq:model_distribution}.\footnote{The data from our experiments are available at \url{https://zenodo.org/record/4551725}}$^,$\footnote{Implementation of Alg. \ref{alg:executionModelSelection} available at \url{https://github.com/alex-mitrevski/explainable-robot-execution-models}}

    \subsection{Object Grasping}

    In the grasping experiment, the robot is placed in front of a table and, for every trial, a single object is placed on the table, which the robot needs to find and grasp. A Faster R-CNN model \cite{ren2015} trained on the YCB subset is used for object detection and recognition.\footnote{Available at \url{https://github.com/b-it-bots/mas_models/tree/master/perception_models/detectors/ycb}}${}^,$\footnote{We used real fruits in the experiments, except for a plastic strawberry; this was due to the fact that real strawberries seemed to be too small for the detector. The model is also unable to recognise the wine glass and the racquetball as such, but recognises them as a chips can and apple respectively; in the generalisation trials with these two objects, we treated them as if they were recognised correctly by the model since this does not affect the results of the experiments.} The grasping action is parameterised by the grasping pose, which is represented by (i) the relative gripper position with respect to the center of the object's bounding box and (ii) the relative wrist orientation with respect to an estimated object orientation (in the range $\left[-\frac{\pi}{2}, \frac{\pi}{2}\right]$). The robot executes a grasp at the selected pose and the outcome is evaluated by a teacher.\footnote{A grasp is considered successful if the object remains in the gripper when the robot retracts the arm back after grasping.} To allow generalisation to differently sized objects, the position parameters are normalised to the range $[-1,1]$ for all axes in the learned model (motivated by \cite{brandl2014}).

    For simplicity of the experimental setup, the robot attempts a sideways grasping strategy for all objects; this also illustrates the limited generalisability between objects more clearly. To estimate the object's position and size before grasping, we find the largest point cloud cluster within the 2D bounding box of the detected object; the center of the cluster is then taken to be the object's position, while the span of the points represents the object's size. To find the planar orientation of the object with respect to the robot, we use a RANSAC-like procedure \cite{fischler1981} that finds the direction of 2D lines fitted to randomly selected points from the object's point cloud; the average of these line orientations is taken to represent the object's orientation.\footnote{We estimate an object's orientation using subsets of the object's point cloud instead of the full point cloud for computational reasons: particularly for larger objects, processing the full point cloud at once takes several seconds, which is not very practical.}

    The relational model of the grasping action is extracted from the following relations (similar to \cite{mitrevski2020}):
    \begin{mdframed}[style=frameStyle]
        \vspace*{-0.25cm}
        \begin{align*}
            \begin{split}
                in\_front\_of_{x,y}(p, B) &\coloneqq p_{x,y} < \min(B_{x,y})\\
                behind_{x,y}(p, B) &\coloneqq p_{x,y} > \max(B_{x,y}) \\
                above(p, B) &\coloneqq p_z > \max(B_z) \\
                below(p, B) &\coloneqq p_z < \min(B_z) \\
                centered_{x,y,z}(p, B) &\coloneqq \left|p_{x,y,z} - \overline{B_{x,y,z}}\right| \leq \epsilon \\
                perpendicular\_to(\theta_o, \theta_p) &\coloneqq |\theta_o - \theta_p| \approx \frac{\pi}{2} \\
                parallel\_to(\theta_o, \theta_p) &\coloneqq |\theta_o - \theta_p| < \theta
            \end{split}
        \end{align*}
    \end{mdframed}
    Here, $\operatorname{rel}_{a_1,...,a_n}$ means that the relation $\operatorname{rel}$ is defined for each of the axes $a_1, ..., a_n$, $p$ is the grasping pose, $B$ is the object's bounding box, $\min/\max$ are the minimum and maximum coordinates along a given axis, $\theta_o$ and $\theta_p$ are the object's planar orientation and the wrist orientation, respectively, $\epsilon$ is a distance threshold set to $5cm$, and $\theta$ is an orientation threshold set to $25^{o}$, which is a conservative value to allow for noisy angle estimates.

    \subsection{Object Stowing}

    For the stowing experiment, the robot is positioned in front of an open drawer and an object is placed in the robot's gripper; the robot needs to throw the object in the drawer in an orderly fashion, namely without damaging the object and so that the drawer can be closed afterwards. As in the grasping case, throws are evaluated by a teacher.\footnote{Throws from large heights are evaluated as unsuccessful for some of the objects, such as the mug, even if the object ends up in the drawer. Similarly, throws that prevent the drawer from closing afterwards or in which the object falls into the drawer after a lucky bounce off the edges are evaluated as unsuccessful.} For repeatability, the object is held by the robot in a similar orientation for all trials; this corresponds to the orientation in which the object is held in the successful grasps of the grasping experiment. Before throwing, the drawer is identified by handle detection (using the same detection model as in \cite{mitrevski2020}); the drawer itself is represented as a box with a known size. The action is parameterised by the throwing pose, which is represented by (i) the relative gripper position with respect to the bounding box of the drawer\footnote{To account for the fact that some part of an object may be extended below the gripper, the average extended length in the successful grasp trials is added to the relative height.} and (ii) the absolute wrist orientation when throwing the object. The position relations used for the grasping model are also used for the relational model of the stowing action; the orientation predicates are used as well, but they evaluate the absolute gripper orientation at which the object is thrown.

    \subsection{Results}

    The number of successful learning trials for each action and training object is shown in Table \ref{tab:successful_learning_grasps_and_throws}.

    \begin{table}[H]
        \centering
        \caption{Successful executions in the data used for learning object-specific execution models (out of 25)}
        \label{tab:successful_learning_grasps_and_throws}
        \begin{tabular}{p{0.28\linewidth} | p{0.28\linewidth} | p{0.28\linewidth}}
            \cellcolor{gray!10!white} Object & \cellcolor{gray!10!white} Grasps & \cellcolor{gray!10!white} Stows \\\hline
            Apple       &  9 & 11 \\\hline
            Chips can   & 11 &  8 \\\hline
            Mug         &  9 &  4 \\\hline
            Sugar box   & 15 &  7 \\\hline
            Tennis ball & 17 & 15 \\\hline
        \end{tabular}
    \end{table}

    As could be expected, several of the learning trials failed due to the fact that execution parameters were selected randomly (position parameters within the object and drawer bounding boxes respectively, throwing height up to around $30cm$ from the top of the drawer, and arbitrary grasping and throwing orientations). Most grasp failures were caused by objects slipping out of the gripper or by the robot attempting to grasp a bit too far from the object. Failed throws were caused either by the object being thrown near the edges of the drawer or by an unreasonably large throwing height.

    The results of the generalisation experiments for the two actions are shown in Table \ref{tab:experimental_results}. We report (i) the size of the object cluster $C_o$ (as we only learned models for five objects, $|C_o|$ varies between $1$ and $2$ for the test objects\footnote{If $C_o$ was not used, all training objects could be considered for generalisation; this increases the number of trials required for identifying an appropriate model and does not scale with the number of objects.}), (ii) the number of attempted models for generalisation (in case of multiple available models), (iii) the object class $o^{*}$ whose model generalises best to each object class $o$ if $\exists o^{*} \in C_o$ for which $P(S=1|o^{*},o) \geq \tau$, and (iv) the total number of successful executions $N^{+}$ over the test trials. During the experiments, we used $\alpha_0 = \beta_0 = 3$ as prior values for the success distribution; we use $\tau = 0.6$ to decide if any of the existing models can be generalised to a test object.\footnote{$\alpha_0 = \beta_0 = 3$ and $\tau = 0.6$ means that at least seven successes in $10$ executions are needed to conclude that a model can be generalised to another object class.}

    \begin{table*}[t]
        \centering
        \scriptsize
        \caption{Generalisation results for the test objects after $10$ executions. We include the size of $C_{o}$, the number of attempted models during generalisation, the object class $o^{*}$ whose model is selected for generalisation (or / in case $P(S=1|\tilde{o},o) < \tau$ $\forall \tilde{o} \in C_{o}$), and the total number of successful executions $N^{+}$ over the $10$ executions.}
        \label{tab:experimental_results}
        \begin{tabular}{l | c | c | c | c | c | c | c | c | c | c | c}
            \hline
            \cellcolor{gray!10!white} Action & \cellcolor{gray!10!white} Object & \cellcolor{gray!10!white} Banana & \cellcolor{gray!10!white} Orange & \cellcolor{gray!10!white} Strawberry & \cellcolor{gray!10!white} Cracker box & \cellcolor{gray!10!white} Can & \cellcolor{gray!10!white} Container & \cellcolor{gray!10!white} Pitcher & \cellcolor{gray!10!white} Wine glass & \cellcolor{gray!10!white} Baseball & \cellcolor{gray!10!white} Racquetball \\\hline
            & \cellcolor{gray!10!white} $|C_o|$       &  1  &   1   &   1   &  2  &     2     &  2  &  2  &  1  &      1      &       1      \\\hline\hline
            \multirow{3}{*}{Grasp}
            & \cellcolor{gray!10!white} \#models      &  1  &   1   &   1   &  1  &     1     &  2  &  2  &  1  &      1      &       1      \\
            & \cellcolor{gray!10!white} $o^{*}$       &  /  & apple & apple &  /  & sugar box &  /  &  /  & mug & tennis ball &  tennis ball \\
            & \cellcolor{gray!10!white} $N^+$         &  4  &   7   &   9   &  5  &     8     &  2  &  1  &  8  &      8      &       7      \\\hline\hline
            \multirow{3}{*}{Stow}
            & \cellcolor{gray!10!white} \#models      &   1   &   1   &   1   &     2     &     1     &     1     &     1     & 1 &      1       &      1      \\
            & \cellcolor{gray!10!white} $o^{*}$       & apple & apple & apple & sugar box & sugar box & chips can & sugar box & / &  tennis ball & tennis ball \\
            & \cellcolor{gray!10!white} $N^+$         &   9   &  10   &   7   &     7     &     9     &     9     &     8     & 1 &     10       &      9      \\\hline
        \end{tabular}
    \end{table*}

    \paragraph{Grasping} As the results of the grasping experiment show, the learned models can be reliably generalised to some of the test objects, but the acquired knowledge is not sufficient in general. As expected, the model of the tennis ball can be generalised to both other balls, but the racquetball was pushed away in a few of the attempts as it is considerably lighter. Similarly, the apple grasping model can be generalised to the orange, although the orange slipped in a few attempts, and surprisingly to the strawberry. As expected, the model cannot be generalised to the banana, for which a top-down grasping strategy would be more suitable (as in Fig. \ref{fig:model_generalisation_over_objects}); some of the banana grasping attempts succeeded as well, but the banana was deformed in all cases due to an inappropriate grasp approach direction.

    The results are more interesting for the containers. Neither of the sugar box and chips can models can be generalised to the cracker box, pitcher, and mustard container. In the case of the cracker box, the sugar box model was successful in half of the attempts, but does not lead to reliable generalisation as its mostly symmetric shape does not provide enough coverage for constraining the grasping orientation, which is important for the elongated cracker box. The mustard container and the pitcher could also not be grasped using the known models due to being quite slippery and also elongated. The tomato can is grasped reliably, but, surprisingly, using the model of the sugar box and not that of the chips can. This is because both the sugar box and the chips can are equally similar to the tomato can based on the ontology and the sugar box model was initially chosen by chance, such that there was no need to attempt the chips can model due to the mostly successful executions using the sugar box model.

    \paragraph{Stowing} According to the results of the stowing experiment, stowing is considerably simpler than grasping, as the learned models can be generalised to most of the test objects. The model of the tennis ball is reliably generalised to the other balls, except for a single failure with the racquetball, which bounced out of the drawer. The model of the apple can generally be reused for the banana and orange; the model can also be reused for the strawberry, but not as successfully, as the strawberry is sometimes thrown from a large height that would damage the fruit.

    As in the grasping case, the container results are more interesting to consider. The sugar box model is reused for both the cracker box and the pitcher, but the execution failed a few times due to the objects being thrown upright, which means that the drawer could not be closed afterwards, or too close to the drawer edges and bouncing out as a result. The sugar box model is also reused for the tomato can; as in the grasping case, the sugar box was chosen by chance initially, such that the box model is quite reliable for the tomato can since the objects have similar masses. The model of the chips can was only reused for the mustard container, initially by chance and then successfully, with only a single failure due to the object being thrown from a large height. Finally, as could be expected, the wine glass requires delicate treatment, so the model of the mug cannot be generalised to it.\footnote{In fact, a few wine glasses were broken during the experiment.}


\section{DISCUSSION AND CONCLUSIONS}
\label{sec:discussion}

    Our method, represented by the notion of what we refer to as a suitability graph, allows a robot to use relations between objects encoded in an ontology as well as its own experiences in order to generalise its action execution knowledge to object classes that have not been manipulated before. In particular, the ontology serves as a prior that guides the generalisation between classes when the robot does not have sufficient experiential information, but the robot's behaviour will be dominated by its own experiences as it interacts with objects throughout its lifetime. As the experiences are treated as annotations over the ontology, the result is a generalisation strategy suitable for explainability and failure analysis. Additionally, due to the probabilistic nature of the model, a robot can determine that its existing knowledge is insufficient for execution in a given context, which enables a form of lifelong acquisition of execution knowledge.

    Our method is based on the assumption that new object classes which have to be manipulated are already present in the robot's ontology, which means that generalisation to objects that are not encoded in the ontology cannot be done directly; for that, it would be necessary to grow the ontology as completely new objects enter the domain, using methods such as \cite{young2016,lim2014}. Related to this and an aspect that we did not address in this paper is that of how to adjust the learned suitabilities if the object cluster is expanded with a new model for a related object class, as this means that there is no generalisation information about that class; future work should investigate strategies to include such models without disrupting the previously learned suitabilities. Another important assumption, which is reflected in the definition of an object cluster, is that only classes that belong to the object's family are included in the cluster, although it might sometimes also be desirable to allow a robot to discover object relations that are not directly encoded in the ontology; one possible strategy for this would be to expand the object cluster by utilising information about object affordances \cite{awaad2014, schoeler2016, wang2014} or object materials \cite{liu2020}. The generalisation between objects obtained using our method is also defined per action; however, similar objects will generally be treated similarly across actions, so already learned relations could potentially be transferred between actions as well, which is an aspect that needs to be investigated in future work. Even though our method is not limited to a specific action, it would be desirable to perform a direct comparison with other existing methods, for instance \cite{mahler2017} for grasping. Finally, in this paper, we used the representation in \cite{mitrevski2020} to model actions; however, it should be noted that the suitability graph is not limited to a specific action representation, which allows different policy models to be combined if desired.


\addtolength{\textheight}{-12cm}   


\section*{ACKNOWLEDGMENT}

We would like to thank Ahmed Abdelrahman and Santosh Thoduka for their comments on an earlier draft of this paper.


\bibliographystyle{IEEEtran}
\bibliography{references}

\begin{thebibliography}{10}
\providecommand{\url}[1]{#1}
\csname url@samestyle\endcsname
\providecommand{\newblock}{\relax}
\providecommand{\bibinfo}[2]{#2}
\providecommand{\BIBentrySTDinterwordspacing}{\spaceskip=0pt\relax}
\providecommand{\BIBentryALTinterwordstretchfactor}{4}
\providecommand{\BIBentryALTinterwordspacing}{\spaceskip=\fontdimen2\font plus
\BIBentryALTinterwordstretchfactor\fontdimen3\font minus
  \fontdimen4\font\relax}
\providecommand{\BIBforeignlanguage}[2]{{%
\expandafter\ifx\csname l@#1\endcsname\relax
\typeout{** WARNING: IEEEtran.bst: No hyphenation pattern has been}%
\typeout{** loaded for the language `#1'. Using the pattern for}%
\typeout{** the default language instead.}%
\else
\language=\csname l@#1\endcsname
\fi
#2}}
\providecommand{\BIBdecl}{\relax}
\BIBdecl

\bibitem{beetz2018}
M.~Beetz, D.~Be{\ss}ler, A.~Haidu, M.~Pomarlan, A.~K. Bozcuo{\v g}lu, and
  G.~Bartels, ``{Know Rob 2.0 - A 2nd Generation Knowledge Processing Framework
  for Cognition-Enabled Robotic Agents},'' in \emph{Proc. IEEE Int. Conf.
  Robotics and Automation (ICRA)}, 2018, pp. 512--519.

\bibitem{awaad2014}
I.~Awaad, G.~K. Kraetzschmar, and J.~Hertzberg, ``{Finding Ways to Get the Job
  Done: An Affordance-Based Approach},'' in \emph{Proc. 24th Int. Conf.
  Planning and Scheduling (ICAPS), Robotics Track}, 2014.

\bibitem{alarcos2019}
A.~Olivares-Alarcos \emph{et~al.}, ``A review and comparison of ontology-based
  approaches to robot autonomy,'' \emph{The Knowledge Engineering Review},
  vol.~34, p. e29, 2019.

\bibitem{paulius2019}
D.~Paulius and Y.~Sun, ``{A Survey of Knowledge Representation in Service
  Robotics},'' \emph{Robotics and Autonomous Systems}, vol. 118, pp. 13--30,
  2019.

\bibitem{schneider2014}
S.~Schneider, N.~Hochgeschwender, and G.~K. Kraetzschmar, ``{Declarative
  Specification of Task-based Grasping with Constraint Validation},'' in
  \emph{Proc. IEEE/RSJ Int. Conf. Intelligent Robots and Systems}, 2014, pp.
  919--926.

\bibitem{mahler2017}
J.~Mahler \emph{et~al.}, ``{Dex-Net 2.0: Deep Learning to Plan Robust Grasps
  with Synthetic Point Clouds and Analytic Grasp Metrics},'' 2017.

\bibitem{shao2020}
L.~Shao, T.~Migimatsu, and J.~Bohg, ``{Learning to Scaffold the Development of
  Robotic Manipulation Skills},'' in \emph{Proc. IEEE Int. Conf. Robotics and
  Automation (ICRA)}, 2020, pp. 5671--5677.

\bibitem{mitrevski2020}
A.~Mitrevski, P.~G. Pl{\"o}ger, and G.~Lakemeyer, ``{Representation and
  Experience-Based Learning of Explainable Models for Robot Action
  Execution},'' in \emph{Proc. IEEE/RSJ Int. Conf. Intelligent Robots and
  Systems (IROS)}, Oct. 2020, pp. 5641--5647.

\bibitem{yamamoto2019}
T.~Yamamoto, K.~Terada, A.~Ochiai, F.~Saito, Y.~Asahara, and K.~Murase,
  ``{Development of Human Support Robot as the research platform of a domestic
  mobile manipulator},'' \emph{ROBOMECH Journal}, vol.~6, pp. 1--15, 2019.

\bibitem{stueber2018}
J.~St{\"u}ber, M.~Kopicki, and C.~Zito, ``{Feature-Based Transfer Learning for
  Robotic Push Manipulation},'' in \emph{Proc. IEEE Int. Conf. Robotics and
  Automation (ICRA)}, 2018, pp. 5643--5650.

\bibitem{liu2020}
W.~Liu, A.~Daruna, and S.~Chernova, ``{CAGE: Context-Aware Grasping Engine},''
  in \emph{Proc. IEEE Int. Conf. Robotics and Automation (ICRA)}, 2020, pp.
  2550--2556.

\bibitem{song2010}
D.~Song, K.~Huebner, V.~Kyrki, and D.~Kragic, ``Learning task constraints for
  robot grasping using graphical models,'' in \emph{Proc. IEEE/RSJ Int. Conf.
  Intelligent Robots and Systems (IROS)}, 2010, pp. 1579--1585.

\bibitem{denninger2018}
M.~Denninger and R.~Triebel, ``{Persistent Anytime Learning of Objects from
  Unseen Classes},'' in \emph{Proc. IEEE/RSJ Int. Conf. Intelligent Robots and
  Systems (IROS)}, 2018, pp. 4075--4082.

\bibitem{young2016}
J.~Young, V.~Basile, L.~Kunze, E.~Cabrio, and N.~Hawes, ``{Towards Lifelong
  Object Learning by Integrating Situated Robot Perception and Semantic Web
  Mining},'' in \emph{Proc. 22nd European Conf. Artificial Intelligence}, 2016,
  pp. 1458--1466.

\bibitem{schoeler2016}
M.~Schoeler and F.~W{\"o}rg{\"o}tter, ``{Bootstrapping the Semantics of Tools:
  Affordance Analysis of Real World Objects on a Per-part Basis},'' \emph{IEEE
  Trans. Cognitive and Developmental Systems}, vol.~8, no.~2, pp. 84--98, June
  2016.

\bibitem{gajewski2019}
P.~Gajewski \emph{et~al.}, ``{Adapting Everyday Manipulation Skills to Varied
  Scenarios},'' in \emph{Proc. IEEE Int. Conf. Robotics and Automation (ICRA)},
  2019, pp. 1345--1351.

\bibitem{bauer2020}
A.~S. Bauer, P.~Schmaus, F.~Stulp, and D.~Leidner, ``{Probabilistic Effect
  Prediction through Semantic Augmentation and Physical Simulation},'' in
  \emph{Proc. IEEE Int. Conf. Robotics and Automation (ICRA)}, 2020, pp.
  9278--9284.

\bibitem{leidner2012}
D.~Leidner, C.~Borst, and G.~Hirzinger, ``{Things are made for what they are:
  Solving manipulation tasks by using functional object classes},'' in
  \emph{12th IEEE-RAS Int. Conf. Humanoid Robots (Humanoids 2012)}, 2012, pp.
  429--435.

\bibitem{sushkov2012}
O.~O. Sushkov and C.~Sammut, ``{Active robot learning of object properties},''
  in \emph{Proc. IEEE/RSJ Int. Conf. Intelligent Robots and Systems (IROS)},
  2012, pp. 2621--2628.

\bibitem{sanan2019}
S.~Sanan, M.~Bretan, and L.~Heck, ``{Learning Object Models For Non-prehensile
  Manipulation},'' in \emph{Proc. IEEE/RSJ Int. Conf. Intelligent Robots and
  Systems (IROS)}, 2019, pp. 4784--4789.

\bibitem{ivaldi2014}
S.~Ivaldi \emph{et~al.}, ``{Object Learning Through Active Exploration},''
  \emph{IEEE Trans. Autonomous Mental Development}, vol.~6, no.~1, pp. 56--72,
  2014.

\bibitem{koller2009}
D.~Koller and N.~Friedman, ``Parameter estimation,'' in \emph{{Probabilistic
  Graphical Models: Principles and Techniques}}.\hskip 1em plus 0.5em minus
  0.4em\relax The MIT Press, 2009, ch.~17, pp. 733--739.

\bibitem{wu1994}
Z.~Wu and M.~Palmer, ``{Verbs Semantics and Lexical Selection},'' in
  \emph{Proc. 32nd Annual Meeting on Association for Computational
  Linguistics}, 1994, pp. 133--138.

\bibitem{calli2015}
B.~Calli, A.~Singh, A.~Walsman, S.~Srinivasa, P.~Abbeel, and A.~M. Dollar,
  ``{The YCB object and Model set: Towards common benchmarks for manipulation
  research},'' in \emph{Int. Conf. Advanced Robotics (ICAR)}, 2015, pp.
  510--517.

\bibitem{ren2015}
S.~Ren, K.~He, R.~Girshick, and J.~Sun, ``{Faster R-CNN: Towards Real-Time
  Object Detection with Region Proposal Networks},'' in \emph{Advances in
  Neural Information Processing Systems 28}, 2015, pp. 91--99.

\bibitem{brandl2014}
S.~Brandl, O.~Kroemer, and J.~Peters, ``{Generalizing Pouring Actions Between
  Objects using Warped Parameters},'' in \emph{14th IEEE-RAS Int. Conf.
  Humanoid Robots (Humanoids)}, Nov. 2014, pp. 616--621.

\bibitem{fischler1981}
M.~A. Fischler and R.~C. Bolles, ``{Random Sample Consensus: A Paradigm for
  Model Fitting with Applications to Image Analysis and Automated
  Cartography},'' \emph{Communications of the ACM}, vol.~24, no.~6, pp.
  381--395, June 1981.

\bibitem{lim2014}
G.~H. Lim \emph{et~al.}, ``{Interactive Teaching and Experience Extraction for
  Learning about Objects and Robot Activities},'' in \emph{23rd IEEE Int. Symp.
  Robot and Human Interactive Communication}, 2014, pp. 153--160.

\bibitem{wang2014}
C.~Wang, K.~V. Hindriks, and R.~Babuska, ``{Effective Transfer Learning of
  Affordances for Household Robots},'' in \emph{4th Int. Conf. Development and
  Learning and Epigenetic Robotics}, 2014, pp. 469--475.

\end{thebibliography}

\end{document}